\documentclass[conference]{IEEEtran}
\usepackage[latin1]{inputenc}

\usepackage[english]{babel}
\usepackage[babel=true]{csquotes}

\usepackage[dvips]{graphicx}

\usepackage{multicol}
\usepackage{subfigure}
\usepackage{amsmath}
\usepackage{amssymb}
\usepackage{amsfonts}
\usepackage{bm}  % Special symbols for sets -- see $\bm{\mathbb{N Z Q R C}}$
\usepackage{oubraces} % To make overlapping under and overbraces

\usepackage{etoolbox}
\let\bbordermatrix\bordermatrix
\patchcmd{\bbordermatrix}{8.75}{4.75}{}{}
\patchcmd{\bbordermatrix}{\left(}{\left[}{}{}
\patchcmd{\bbordermatrix}{\right)}{\right]}{}{}
\usepackage{url}

\usepackage{savefnmark} % to get pointer to same footnote (if necessary) - Added by Jean
\usepackage{eurosym}

\usepackage{stackrel} % To stack symbols, see https://tex.stackexchange.com/questions/39225/differences-between-stackrel-and-stackbin

\usepackage{color}  % emphasize comments and use also convenient \sout{xxx} \uline{xxx} latex command as well
\usepackage{relsize} % Allow to enlarge math symbols in equations (used in Shafer's belief inequality)

\usepackage{cite} % Loading the cite package will result in citation numbers being automatically sorted and properly "ranged". i.e., [1], [9], [2], [7], [5], [6] (without using cite.sty) will become: [1], [2], [5]--[7], [9] (using cite.sty)
\usepackage{stfloats}  % Gives LaTeX2e the ability to do double column
 \usepackage[babel=true]{csquotes}

\ifCLASSINFOpdf
\else
\fi
\usepackage{url}
\begin{document}

%\title{Total Belief and Total Plausibility Theorems\\

%\pagenumbering{arabic}
%\pagestyle{empty}

 \onecolumn
 
\title{Distances Between Partial Preference Orderings}
 
%  \title{Uncertainty Representation Using Belief Functions}

%\author{\IEEEauthorblockN{Jean~Dezert\IEEEauthorrefmark{a}, Andrii~Shekhovtsov\IEEEauthorrefmark{b}, Wojciech~Sa\l{}abun\IEEEauthorrefmark{b} \\[7pt]
%\IEEEauthorblockA{\IEEEauthorrefmark{a}The French Aerospace Lab, Palaiseau, France}
%\IEEEauthorblockA{\IEEEauthorrefmark{b}National Telecommunication Institute, Warsaw, Poland}
%Emails: \url{jean.dezert@onera.fr},  \url{andrii-shekhovtsov@il-pib.pl},  \url{w.salabun@il-pib.pl}
%}}

\author{
\IEEEauthorblockN{Jean~Dezert$^a$, Andrii~Shekhovtsov$^b$, Wojciech~Sa\l{}abun$^b$ \\[5pt]
$^a$The French Aerospace Lab, Palaiseau, France.\\
$^b$National Telecommunication Institute, Warsaw, Poland.\\
Emails: \url{jean.dezert@onera.fr},  \url{andrii-shekhovtsov@il-pib.pl},  \url{w.salabun@il-pib.pl}
}}

\maketitle

% Page numbering (if needed)
%=====================
% Uncomment these two lines below to display page numbers
\thispagestyle{plain}
\pagestyle{plain}


\begin{abstract}
This paper proposes to establish the distance between partial preference orderings based on two very different approaches. The first approach corresponds to the brute force method based on combinatorics. It generates all possible complete preference orderings compatible with the partial preference orderings and calculates the Frobenius distance between all fully compatible preference orderings. Unfortunately, this first method is not very efficient in solving high-dimensional problems because of its big combinatorial complexity. That is why we propose to circumvent this problem by using a second approach based on belief functions, which can adequately model the missing information of partial preference orderings. This second approach to the calculation of distance does not suffer from combinatorial complexity limitation. We show through simple examples how these two theoretical methods work.
\end{abstract}

\medskip



% \fi
%\IEEEpeerreviewmaketitle

%===================================
%===================================
\section{Introduction}
%===================================
%===================================

In operational research and multi-criteria decision-making (MCDM) problems, we are often faced with dealing with information expressed as preference orderings (PO), especially if human experts provide their opinions about the problem under concern only as preferences between several objects (or alternatives) and express qualitatively or quantitatively. The PO given by the source of information are not always total preference orderings (TPOs) (i.e., full or complete PO) but only defined partially, and we have to deal efficiently with this type of partial preference orderings (PPOs) to infer some important decision to make.   

There exist several works on metrics and near metrics (i.e., similarity measures) between ranked data, and ranking aggregation is problematic, see \cite{Critchlow1985,Fagin2004,Fagin2006,draws2023}. Most of these works are based on Spearman's footrule \cite{diaconis1977spearman} and Kendall's correlation coefficient  \cite{Bansal2009,Albano2021}, and they fail to take into account element relevance and positional information \cite{Kumar2010}. An appealing axiomatic approach to define the distance between TPO had been proposed in the sixties by Kemeny \cite{Kemeny1962,Albano2021} and is often used in applications. Recently, we have proved in \cite{Frobenius2023} that Kemeny's distance is not the unique distance satisfying Kemeny's axioms, and the Frobenius distance can be used instead for measuring the distance between two TPOs. This paper extends our previous work and shows how it is possible to exploit this Frobenius distance to measure the distance between two PPOs without involving combinatorics.

Before presenting in detail the theoretical solutions we propose, we recall briefly the basics about the comparison of objects in Section \ref{Sec2}, Frobenius distance between total preference orderings in Section \ref{Sec3}, and belief functions in Section~\ref{Sec4}. Then, we present the brute force method (BFM) based on combinatorics in Section \ref{Sec5} with examples for calculating the distance between PPOs. To circumvent the complexity limitations of BFM in Section \ref{Sec6}, we propose two methods of distance calculation based on belief functions, and we compare them with BFM in some examples. Section \ref{Sec7} concludes this paper with some perspectives on MCDM and information fusion applications.

%===================================
%===================================
\section{Comparison of two objects}
%===================================
%===================================
\label{Sec2}

Let's consider two objects\footnote{The {\it{object}} terminology used here is a generic term representing either physical objects, hypotheses, alternatives, etc, depending on the context of the problem.} $A$ and $B$, with their score denoted by $c(A)$ and $c(B)$ based on some chosen criteria of evaluation $c(.)$.
The score value can be quantitative (i.e., a number) or qualitative\footnote{If the score is qualitative, we assume that a strict order of qualitative label exists and is known, say if we consider labels {\it{Very Good}}, {\it{Good}} and {\it{Bad}}, then we consider that {\it{Very Good}} is better than  {\it{Good}}, which is better than  {\it{Bad}}, i.e. we have the order $ \text{\it{Very Good}} >  \text{\it{Good}} > \text{\it{Bad}}$.} (i.e. a label). Based on the score values and adopting the convention \enquote{higher score is better}, in the classical preference model, we assume that we can always compare two objects and express the preference ordering between these two objects as follows:
\begin{itemize}
\item $A$ preferred to $B$ for sure is denoted $A\succ B$ where $A\succ B \Leftrightarrow c(A)>c(B)$, which is equivalent to $B\prec A$ where $B\prec A \Leftrightarrow c(B)<c(A)$.
\item $B$ preferred to $A$ for sure is denoted $B\succ A$ where $B\succ A \Leftrightarrow c(B)>c(A)$, which is equivalent to $A\prec B$ where $A\prec B \Leftrightarrow c(A)<c(B)$.
\item Indifference: $A$ is equivalent to $B$ for sure is denoted $A\equiv B$, and one has $A\equiv B \Leftrightarrow c(A)=c(B)$, 
which is equivalent to $B\equiv A$ where $B\equiv A \Leftrightarrow c(B)=c(A)$.
$A$ is equivalent to $B$ for sure means that we know that the inequalities $c(A)>c(B)$ and $c(A)<c(B)$ are not valid. Hence, we are indifferent to $A$ and to $B$.
\end{itemize}

Note that the preferences are not necessarily based on some objective (quantitative or qualitative) criterion $c(.)$,  but they can sometimes be based on some personal intuition or expert knowledge related to the problem under consideration, especially if they come from human sources.

%===================================
\section{Distance between two TPOs}
%===================================
\label{Sec3}

We consider here a set $X$ of $n\geq 2$ objects that are all ranked by two sources of information, and we denote $\text{Pref}_1$ and $\text{Pref}_2$ each total\footnote{A preference ordering is said {\it{total}}  or {\it{complete}} or {\it{full}} if all available objects under concern appear in the preference order. If some objects do not appear in the preference order, we say it is a partial preference ordering (PPO).} preference ordering (TPOs) of objects given by the sources. For instance, if we consider a set of three objects $X=\{x_1=A,x_2=B,x_3=C\}$, then we may have the preference $\text{Pref}_1$ given by the source 1, and the preference $\text{Pref}_2$ by the source 2, with the following TPOs ${\text{Pref}_1\triangleq A\succ B \succ C}$ and ${\text{Pref}_2\triangleq B\succ C \succ  A}$. 
% The preference symbol $\succ$ means that if $A\succ B$, then $A$ is preferred to $B$; also, if $B \succ C$ means that $B$ is preferred to $C$, etc.

In this section, we explain how to measure the distance between two TPOs ${\text{Pref}_1}$ and ${\text{Pref}_1}$ in general case. For this, we use the Frobenius distance between two total orderings (i.e. preference orderings including eventual ties) proposed in \cite{Frobenius2023} as a serious alternative to classical Kemeny's distance \cite{Kemeny1962}, and to Spearman's footrule distance (i.e. $F$-distance) \cite{diaconis1977spearman}. The advantage of Frobenius distance is that it is a true metric, and it is quite simple to calculate. Contrary to the $F$-distance, the Frobenius distance satisfies the essential invariance under the indexing principle \cite{Frobenius2023}. Moreover, the Frobenius distance also satisfies all of Kemeny's axioms. Hence, the well-known Kemeny's distance is not unique, and we have shown why Frobenius's distance is appealing with respect to Kemeny's distance through simple examples in \cite{Frobenius2023}.

%===================================
\subsection{Definition}
%===================================

Frobenius distance between two TPOs (i.e. orderings) of $N$ objects is done by calculating at first the ${N\times N}$ pairwise Preference-Score Matrix (PSM) based on the ordering given by each expert. By convention, the row index $i$ of the PSM corresponds to the index of elements $x_i$ on the left side of preference ordering $x_i \succ x_j$, and the column index $j$ of the PSM corresponds to the index of the element $x_j$ on the right side of preference ordering ${x_i \succ x_j}$. Hence we denote a pairwise Preference-Score Matrix $\bm{M}(X)=[{M}(i,j)]$ where its components ${{M}(i,j)}$ for $i,j=1,2,\ldots,N$ are defined as
\begin{equation}
\bm{M}(i,j)=
\begin{cases}
1,  &\quad \text{if $x_i\succ x_j$},\\
-1, &\quad \text{if $x_i\prec x_j$},\\
0, &\quad \text{if $x_i=x_j$}.
\end{cases}
\label{eqPSM}
\end{equation}

Note that all components ${M}(i,i)$ ($i=1,2,\ldots,N$) of the main diagonal of the matrix $\bm{M}$ are always equal to zero. Also, PSM is always an anti-symmetrical matrix by construction because the preference ${x_i\succ x_j}$ is equivalent to the preference ${x_j \prec x_i}$. Hence if $x_i\succ x_j$ is true which means ${M}(i,j)=1$ then necessarily ${x_j\succ x_i}$ is false which means that ${x_j\prec x_i}$ is true and thus ${M}(j,i)=-1$, and the other way around. Consequently, ${\bm{M}(X)^T=-\bm{M}(X)}$, and $\text{Tr}(\bm{M}(X))=0$. 

The distance between two TPOs $\text{Pref}_1$ and $\text{Pref}_2$ is defined by the Frobenius distance\footnote{We use the subscript F in our notation to refer to Frobenius.} as follows \cite{Frobenius2023}:
\begin{equation}
d_F(\bm{M}_1,\bm{M}_2)={||\bm{M}_1 - \bm{M}_2 ||}_F,
\label{FrobeniusDistance}
\end{equation}
where $||\bm{M}||_F$ is the Frobenius norm of a square matrix $\bm{M}=[M(i,j),i,j=1,\ldots,N]$ defined by \cite{Horn1990,Golub1996} 
\begin{equation}
||\bm{M}||_F=\sqrt{ \sum_{i=1}^n \sum_{j=1}^n |M(i,j)|^2}=\sqrt{\text{Tr}(\bm{M}^T\bm{M})}, 
\end{equation}
and where $\bm{M}^T$  is the transpose of the matrix $\bm{M}$, and $\text{Tr}(.)$ is the trace operator for matrix.

Frobenius distance is usually normalized in $[0,1]$ interval by dividing the value $d_F(\bm{M}_1,\bm{M}_2)$ by the maximal distance value $d_F^{\max}$ calculated by considering two TPOs in full contradiction (i.e. a preference and its opposite defined by reversing the preference order).
For instance, if a preference ordering is $\text{Pref}=A\succ B \succ C \succ D$, its opposite is $\neg \text{Pref}=A\prec B \prec C \prec D = D\succ C \succ B \succ A$.

The PSM established from formula \eqref{eqPSM} proposed by Kemeny in \cite{Kemeny1962} may appear quite ad-hoc and arbitrary, and some authors prefer another PSM definition like the $\bm{M}'$ matrix based on the formula \eqref{eqPSM2} below. In fact what really matters in the construction of PSM is to keep the principle of what PSM represents\footnote{that is, if $x_i\succ x_j$ the PSM value $\bm{M}(i,j)$ must be highest; if $x_i\prec x_j$ the PSM value must be lowest; and if $x_i=x_j$ the PSM value $\bm{M}(i,j)$ must be in-between highest and lowest values.}.
\begin{equation}
\bm{M}'(i,j)=
\begin{cases}
1,  &\quad \text{if $x_i\succ x_j$},\\
0, &\quad \text{if $x_i\prec x_j$},\\
0.5, &\quad \text{if $x_i=x_j$}.
\end{cases}
\label{eqPSM2}
\end{equation}

Even if the PSM definition changes the Frobenius distance value, it will not impact the normalized distance value (see the next example). That is why it is preferable to work with normalized distances in applications when comparing different distances for a given purpose. Therefore, we work with normalized distances in this paper. 

%===================================
\subsection{Example 1}
%===================================
\label{Example1}
\noindent
Consider the set $X=\{x_1=A, x_2=B, x_3=C\}$ of three objects $A$, $B$ and $C$, and the two total preference orderings 
$$
\begin{cases}
{\text{Pref}_1\triangleq B \succ A \succ C},\\
{\text{Pref}_2\triangleq B\succ C \succ A}.
\end{cases}
$$
Based on \eqref{eqPSM}, we obtain the PSM matrices
{\small{
\begin{equation*}
{\bm{M}_1}  =
\begin{bmatrix}
0 & -1 & 1  \cr
1 & 0 & 1 \cr
 -1 & - 1 & 0 \cr			
\end{bmatrix}, \ \text{and}\ 
%\label{eqMX1}
%\text{and}	\ 			
{\bm{M}_2}  =
\begin{bmatrix}
 0 & -1 & -1  \cr
1 & 0 & 1 \cr
1 & - 1 & 0  \cr		
\end{bmatrix}.	
\label{eqMX2}						
\end{equation*}
}}
Because 
{\small{
\begin{equation}
(\bm{M}_1 -\bm{M}_2)^T (\bm{M}_1 -\bm{M}_2) =
\begin{bmatrix}
    4  &   0  &   0  \\
     0  &   0 &   0  \\
     0  &   0  &   4   \\
     \end{bmatrix}, 
\end{equation}
}}
one has $\text{Tr}(\bm{M}_1 -\bm{M}_2)^T (\bm{M}_1 -\bm{M}_2) = 8$.

Applying \eqref{FrobeniusDistance} with PSMs $\bm{M}_1$ and $\bm{M}_2$ the Frobenius distance between $\text{Pref}_1$ and $\text{Pref}_2$ is equal to
$$ d_F(\bm{M}_1,\bm{M}_2)=\sqrt{8}\approx 2.8284.$$

%$d_F(\bm{M}_1,\bm{M}_2)$ by its maximal value $d_F^\max$ calculated by considering two total preferences orderings in full contradiction (i.e. a preference and its opposite defined by reversing the preference order). For instance, if a preference ordering is $\text{Pref}=A\succ B \succ C \succ D$, its opposite is $\neg \text{Pref}=A\prec B \prec C \prec D = D\succ C \succ B \succ A$.

The maximal value $d_F^{\max}$ of $d_F(\bm{M}_1,\bm{M}_2)$ is calculated in considering two TPOs in full contradiction (i.e., a preference and its opposite defined by reversing the preference order). For instance, in considering $\text{Pref}=A\succ B \succ C$, and its opposite $\neg \text{Pref}=A\prec B \prec C = C \succ B \succ A$ and building PSM by \eqref{eqPSM} and applying \eqref{FrobeniusDistance}. In our example we obtain,  $ d_F^{\max} = \sqrt{24}\approx 4.8990$.

Therefore, the normalized Frobenius distance $\tilde{d}_F$ between ${\text{Pref}_1}$ and ${\text{Pref}_2}$ is given by
$$\tilde{d}_F(\bm{M}_1,\bm{M}_2)=\sqrt{8}/\sqrt{24}=1/\sqrt{3} \approx 0.5774.$$

Using the alternative PSM definition \eqref{eqPSM2}, we obtain the PSM matrices
{\small{
\begin{equation*}
{\bm{M}'_1}  =
\begin{bmatrix}
0.5 & 0 & 1 \cr
1 & 0.5 & 1  \cr
 0 & 0 & 0.5  \cr		
\end{bmatrix}, \ \text{and}\ 
%\label{eqMX1}
%\text{and}	\ 			
{\bm{M}'_2}  =
\begin{bmatrix}
 0.5 & 0 & 0  \cr
1 & 0.5 & 1  \cr
1 & 0 & 0.5  \cr		
\end{bmatrix}.	
\label{eqMX2prime}						
\end{equation*}
}}
Because 
{\small{
\begin{equation}
(\bm{M}'_1 -\bm{M}'_2)^T (\bm{M}'_1 -\bm{M}'_2) =
\begin{bmatrix}
     1  &   0   &  0  \\
     0  &   0 &    0  \\
     0  &   0  &   1  \\
     \end{bmatrix}, 
\end{equation}
}}
one has $\text{Tr}(\bm{M}'_1 -\bm{M}'_2)^T (\bm{M}'_1 -\bm{M}'_2) = 2$.

Applying \eqref{FrobeniusDistance} with PSMs given in \eqref{eqMX2prime}, the Frobenius distance between $\text{Pref}_1$ and $\text{Pref}_2$ is equal to
$$d_F(\bm{M}'_1,\bm{M}'_2)=\sqrt{2}\approx 1.4142.$$

The maximal value $d_F^{\max}$ of $d_F(\bm{M}'_1,\bm{M}'_2)$ is calculated in considering two TPOs in full contradiction. In our example we have $ d_F^{\max} = \sqrt{6}=\sqrt{2}\cdot \sqrt{3}$.

Therefore, the normalized Frobenius distance between ${\text{Pref}_1}$ and ${\text{Pref}_2}$ based on alternative PSM definition \eqref{eqPSM2}, is given by
$$\tilde{d}_F(\bm{M}'_1,\bm{M}'_2)=\sqrt{2}/\sqrt{6}=1/\sqrt{3} \approx  0.5774.$$

So, we see that the choice of PSM definition (either based on \eqref{eqPSM} or on \eqref{eqPSM2}) has no impact on the normalized Frobenius distance between preference orderings.

In our example, because the normalized distance is more close to one than to zero, we can legitimately assert that the preference ${\text{Pref}_2}$ is actually more dissimilar to ${\text{Pref}_1}$ than similar to ${\text{Pref}_1}$, and vice-versa.

%================================
\section{Belief functions}
%================================
\label{Sec4}

The belief functions (BF) were introduced by Shafer \cite{Shafer1976} for modeling epistemic uncertainty, reasoning about uncertainty, and combining distinct sources of evidence. The answer to the problem under concern is assumed to belong to a known finite discrete frame of discernment (FoD) $\Theta=\{\theta_1,\ldots,\theta_N\}$ where all elements (i.e., members) of $\Theta$ are exhaustive and exclusive. The set of all subsets of $\Theta$ (including empty set $\emptyset$, and $\Theta$) is the power-set of $\Theta$ denoted by $2^\Theta$. The number of elements (i.e., the cardinality) of the power set is $2^{|\Theta |}$. A (normalized) basic belief assignment (BBA) associated with a given source of evidence is a mapping $m^\Theta(\cdot):2^\Theta \to [0,1]$ such that\footnote{In Shafer's theory of BFs, we work with a closed FoD and the mass of the empty set must always be equal to zero.} $m^\Theta(\emptyset)=0$ and $\sum_{X\in 2^\Theta} m^\Theta(X) = 1$. A BBA $m^\Theta(\cdot)$ characterizes a source of evidence related to a FoD $\Theta$. For notation shorthand, we can omit the superscript $\Theta$ in $m^\Theta(\cdot)$ notation if there is no ambiguity on the FoD we work with. The quantity $m(X)$ is called the mass of belief of $X$. $X\in 2^\Theta$ is called a focal element (FE) of $m(\cdot)$ if $m(X)>0$. The set of all focal elements of $m(\cdot)$ is denoted 
by $\mathcal{F}_{\Theta}(m)\triangleq \{X\in 2^\Theta | m(X)>0 \}$. The belief and the plausibility of $X$ are respectively defined for any $X\in 2^\Theta$ by \cite{Shafer1976}.

{\small{\begin{equation}
Bel(X) = \sum_{Y\in 2^\Theta | Y\subseteq X} m(Y)\,,
\label{eqBel}
\end{equation}
}}
{\small{
\begin{equation}
Pl(X) =  \sum_{Y\in 2^\Theta | X\cap Y\neq \emptyset} m(Y)=1-\text{Bel}(\bar{X})\,,
\label{eqPl}
\end{equation}
}}
\noindent
where ${\bar{X}\triangleq\Theta\setminus\{X\}}$ is the complement of $X$ in $\Theta$\,.

One has always ${0\leq Bel(X)\leq Pl(X) \leq 1}$, see \cite{Shafer1976}. For ${X=\emptyset}$, ${Bel(\emptyset)=Pl(\emptyset)=0}$, and for ${X=\Theta}$ one has ${Bel(\Theta)=Pl(\Theta)=1}$. $Bel(X)$ and $Pl(X)$ are often interpreted as the lower and upper bounds of unknown probability $P(X)$ of $X$, that is ${Bel(X) \leq P(X) \leq Pl(X)}$. To quantify the uncertainty (i.e. the imprecision) of ${P(X)\in [Bel(X), Pl(X)]}$, we use $u(X)\in [0,1]$ defined by

{\small{
\begin{equation}
u(X)\triangleq Pl(X)-Bel(X)\,.
\label{defuX}
\end{equation}
}}
If all focal elements of $m(\cdot)$ are singletons of $2^\Theta$ the BBA $m(\cdot)$ is a Bayesian BBA because ${\forall X\in2^\Theta}$ one has ${Bel(X)=Pl(X)=P(X)}$ and $u(X)=0$. Hence, the belief and plausibility of $X$ coincide with a probability measure $P(X)$ defined on the FoD $\Theta$. The vacuous BBA characterizing a totally ignorant source of evidence is defined by ${m_v(X)=1}$ for ${X=\Theta}$, and ${m_v(X)=0}$ for all ${X\in 2^\Theta}$ different of $\Theta$.
This very particular BBA is very important to model missing information in the PPOs as it will be shown in Section \ref{Sec6}.

%=======================================================
\section{Distance between two PPOs by BFM}
%=======================================================
\label{Sec5}

In practice, the calculation of the distance between PPOs is of main importance because TPOs are not always available to calculate directly their proximity by the Frobenius distance presented in Section \ref{Sec3}. The main question is how to calculate the distance between two POs when at least one is partially known. From the theoretical standpoint, the combinatorial solution based on the brute force method (BFM) seems appealing because it will enumerate all compatible TPOs (i.e., CTPOs) with the given PPOs, and it computes the Frobenius distances between all CTPOs pairs and makes a decision from of all them based on different decisional attitudes. This section presents in detail the BFM.

\subsection{The brute force method (BFM)}
%===============================

The BFM consists of generating all the TPOs of the objects and then selecting all of them that are compatible with each PPO (called CTPOs) and calculating their corresponding Frobenius distances. As the final result for the normalized distance, we take either:

$\bullet$ $\tilde{d}^{\text{Optim}}$: normalized distance based on the optimistic decision-making attitude (i.e., the minimum of  Frobenius distance among all possible normalized distances of CTPOs);

$\bullet$ $\tilde{d}^{\text{Pessim}}$: normalized distance based on the pessimistic decision-making attitude (i.e., the maximum Frobenius distance among all possible normalized distances of CTPOs);

$\bullet$ $\tilde{d}^{\text{Aver}}$: normalized distance based on the averaged decision-making attitude of all possible normalized distances of CTPOs;

$\bullet$ $\tilde{d}^{\text{Hurwicz}}$: normalized distance based on Hurwicz attitude \cite{Hurwicz1951,Hurwicz1952} which is a weighted balance between the min and max decision-making attitudes.

This BFM approach is acceptable in theory, but unfortunately, it can quickly become intractable because of the high combinatorics involved, even with low-dimension problems. The following simple example shows how to apply BFM to calculate distances between PPOs. This example has been chosen for its simplicity and because the underlying combinatorics is limited. So we can include all the details of the computations. The reader can easily realize how the combinatorics of the BFM approach become intractable if more difficult (bigger) examples are addressed. That is why we consider preference orderings between only three objects: $A$, $B$, and $C$. In this case, only thirteen possible TPOs (including ties) exist. They are listed in the Table \ref{TableTPO}.
\begin{table}[!h]
\begin{center}
\caption{All Possible Total Preference Orderings (TPOs) for 3 objects.\label{TableTPO}}
\begin{tabular}{|c|c|c|}
\hline
$ A\succ B \succ C$ & $ B\succ C \succ A$    & $ A\succ (B \equiv C)$ \\
$ A\succ C \succ B$ & $ C\succ A \succ B$    & $ B\succ (A \equiv C)$ \\
$ B\succ A \succ C$ & $ C\succ B \succ A$    & $ C\succ (A \equiv B)$ \\
$ (A\equiv B) \succ C)$ &  $ (A\equiv C) \succ B)$ & $ (B\equiv C) \succ A)$\\
& $ (A\equiv B\equiv C)$ & \\
\hline
\end{tabular}
\end{center}
\end{table}

The maximum Frobenius distance between two TPOs listed in Table \ref{TableTPO} is equal to $d_F^{\max}=\sqrt{24}\approx 4.8990$.

\subsection{Example 2} 
%=================
\label{Example 2}

Consider the objects $\{A,B,C\}$ and the two PPOs given by $ \text{Pref}_1 = C\succ A$ and $\text{Pref}_2 = A\succ B$. 
%\begin{align*}
%& \text{Pref}_1 = C\succ A\\
%& \text{Pref}_2 = A\succ B 
%\end{align*}
$\text{Pref}_1$ and $\text{Pref}_2$ are both PPOs because they include only two objects among three. For applying BFM, we must consider all TPOs listed in Table \ref{TableEx89} that are compatible with $ \text{Pref}_1$ on one hand and compatible with $ \text{Pref}_2$ on the other hand. By examining all possible TPOs generated by brute force combinatorics (see Table \ref{TableTPO}), we observe that we have only ${N_1^{\text{CTPO}}=5}$ CPTOs with $\text{Pref}_1$ which are listed in Table \ref{TableEx89}, and ${N_2^{\text{CTPO}}=5}$ CTPOs with $\text{Pref}_2$ as listed in Table \ref{TableEx89}.

\begin{table}[!h]
\centering
\caption{CTPOs $\text{Pref}_1^i$ and $\text{Pref}_2^j$ for $\text{Pref}_1 = (C\succ A)$ and $\text{Pref}_2 = (A\succ B)$.}
\label{TableEx89}
\begin{tabular}{|c|c|c|c|c|}
\cline{1-2} \cline{4-5}
$\text{Pref}_1^{1}$ & $ B\succ C \succ A$    &  & $\text{Pref}_2^{1}$ & $ A\succ B \succ C$     \\
$\text{Pref}_1^{2}$ & $ C\succ A \succ B$    &  & $\text{Pref}_2^{2}$ & $ A\succ C \succ B$     \\
$\text{Pref}_1^{3}$ & $ C\succ B \succ A$    &  & $\text{Pref}_2^{3}$ & $ C\succ A \succ B$     \\
$\text{Pref}_1^{4}$ & $ C\succ (A \equiv B)$ &  & $\text{Pref}_2^{4}$ & $ A\succ (B \equiv C)$  \\
$\text{Pref}_1^{5}$ & $ (B\equiv C)\succ A$  &  & $\text{Pref}_2^{5}$ & $ (A\equiv C) \succ B $ \\ \cline{1-2} \cline{4-5} 
\end{tabular}
\end{table}

The PSM $\bm{M}_1^k$ ($k=1,\ldots,N_1^{\text{CTPO}}$) corresponding to CTPOs of Table \ref{TableEx89} are 
%$$
%\footnotesize{
%\bm{M}_1^1  =
%\begin{bmatrix}
%0  & -1 & -1 \cr
%1 & 0 & 1 \cr
% 1 & - 1 & 0\cr	
%\end{bmatrix},
%\bm{M}_1^2  =
%\begin{bmatrix}
%0  & 1 & -1 \cr
%-1 & 0 & -1 \cr
% 1 & 1 & 0\cr	
%\end{bmatrix},
%}
%$$
%
%$$
%\footnotesize{
%\bm{M}_1^3  =
%\begin{bmatrix}
%0  & -1 & -1 \cr
%1 & 0 & -1 \cr
% 1 & 1 & 0\cr	
%\end{bmatrix},
%\bm{M}_1^4  =
%\begin{bmatrix}
%0  & 0 & -1 \cr
%0 & 0 & -1 \cr
% 1 & 1 & 0\cr	
%\end{bmatrix},
%}
%$$
%
%$$
%\footnotesize{
%\bm{M}_1^5  =
%\begin{bmatrix}
%0  & -1 & -1 \cr
%1 & 0 & 0 \cr
% 1 & 0 & 0\cr	
%\end{bmatrix}.
%}
%$$

$$
\footnotesize{
\bm{M}_1^1  =
\begin{bmatrix}
0  & -1 & -1 \cr
1 & 0 & 1 \cr
 1 & - 1 & 0\cr	
\end{bmatrix},
\bm{M}_1^2  =
\begin{bmatrix}
0  & 1 & -1 \cr
-1 & 0 & -1 \cr
 1 & 1 & 0\cr	
\end{bmatrix},
\bm{M}_1^3  =
\begin{bmatrix}
0  & -1 & -1 \cr
1 & 0 & -1 \cr
 1 & 1 & 0\cr	
\end{bmatrix},
\bm{M}_1^4  =
\begin{bmatrix}
0  & 0 & -1 \cr
0 & 0 & -1 \cr
 1 & 1 & 0\cr	
\end{bmatrix},
\bm{M}_1^5  =
\begin{bmatrix}
0  & -1 & -1 \cr
1 & 0 & 0 \cr
 1 & 0 & 0\cr	
\end{bmatrix}.
}
$$

The PSM $\bm{M}_2^k$ ($k=1,\ldots,N_2^{\text{CTPO}}$) corresponding to CTPOs of Table \ref{TableEx89} are 

%$$
%\footnotesize{\bm{M}_2^1  =
%\begin{bmatrix}
%0  & 1 & 1 \cr
%-1 & 0 & 1 \cr
% -1 & - 1 & 0\cr	
%\end{bmatrix},
%\bm{M}_2^2  =
%\begin{bmatrix}
%0  & 1 & 1 \cr
%-1 & 0 & -1 \cr
% -1 & 1 & 0\cr	
%\end{bmatrix},
%}
%$$
%$$
%\footnotesize{
%\bm{M}_2^3  =
%\begin{bmatrix}
%0  & 1 & -1 \cr
%-1 & 0 & -1 \cr
% 1 & 1 & 0\cr	
%\end{bmatrix},
%\bm{M}_2^4  =
%\begin{bmatrix}
%0  & 1 & 1 \cr
%-1 & 0 & 0 \cr
% -1 & 0 & 0\cr	
%\end{bmatrix},
%}
%$$
%$$
%\footnotesize{
%\bm{M}_2^5  =
%\begin{bmatrix}
%0  & 1 & 0 \cr
%-1 & 0 & -1 \cr
% 0 & 1 & 0\cr	
%\end{bmatrix}.
%}
%$$

$$
\footnotesize{\bm{M}_2^1  =
\begin{bmatrix}
0  & 1 & 1 \cr
-1 & 0 & 1 \cr
 -1 & - 1 & 0\cr	
\end{bmatrix},
\bm{M}_2^2  =
\begin{bmatrix}
0  & 1 & 1 \cr
-1 & 0 & -1 \cr
 -1 & 1 & 0\cr	
\end{bmatrix},
\bm{M}_2^3  =
\begin{bmatrix}
0  & 1 & -1 \cr
-1 & 0 & -1 \cr
 1 & 1 & 0\cr	
\end{bmatrix},
\bm{M}_2^4  =
\begin{bmatrix}
0  & 1 & 1 \cr
-1 & 0 & 0 \cr
 -1 & 0 & 0\cr	
\end{bmatrix},
\bm{M}_2^5  =
\begin{bmatrix}
0  & 1 & 0 \cr
-1 & 0 & -1 \cr
 0 & 1 & 0\cr	
\end{bmatrix}.
}
$$

To calculate the normalized distance between the PPOs ${\text{Pref}_1}$ and ${\text{Pref}_2}$ for some chosen decision-making attitude, we must consider all possible pairs of PSM ${(\bm{M}_1^i,\bm{M}_2^j)}$ for ${i=1,\ldots,N_1^{\text{CTPO}} }$ and ${j=1,\ldots,N_2^{\text{CTPO}} }$, and calculate their normalized Frobenius distance ${\tilde{d}_F(\bm{M}_1^i,\bm{M}_2^j)}$. Because we have ${N_1^{\text{CTPO}}=5}$ and ${N_2^{\text{CTPO}}=5}$, we have ${5\times 5=25}$ Frobenius distances to calculate. All the possible normalized distances for PSM pairs ${(\bm{M}_1^i,\bm{M}_2^j)}$ are listed in the Table \ref{CTPOpairsEx2}.

\begin{table}[!h]
\caption{Normalized Frobenius Distances for all CTPOs pairs. \label{CTPOpairsEx2}}
\begin{center}
\begin{tabular}{|c|c|c|c|c|c|}
\hline
$\tilde{d}_F(\bm{M}_1^i,\bm{M}_2^j)$ & $\bm{M}_2^1$ & $\bm{M}_2^2$ & $\bm{M}_2^3$ & $\bm{M}_2^4$ & $\bm{M}_2^5$\\
\hline
$\bm{M}_1^1$ & $\frac{\sqrt{16}}{\sqrt{24}}$ & $\frac{\sqrt{24}}{\sqrt{24}}$ & $\frac{\sqrt{16}}{\sqrt{24}}$ &  $\frac{\sqrt{18}}{\sqrt{24}}$ &  $\frac{\sqrt{18}}{\sqrt{24}}$ \\
$\bm{M}_1^2$ & $\frac{\sqrt{16}}{\sqrt{24}}$ & $\frac{\sqrt{8}}{\sqrt{24}}$   & $\frac{\sqrt{0}}{\sqrt{24}}$  & $\frac{\sqrt{10}}{\sqrt{24}}$ & $\frac{\sqrt{2}}{\sqrt{24}}$\\
$\bm{M}_1^3$ & $\frac{\sqrt{24}}{\sqrt{24}}$ & $\frac{\sqrt{16}}{\sqrt{24}}$ & $\frac{\sqrt{8}}{\sqrt{24}}$  &  $\frac{\sqrt{18}}{\sqrt{24}}$ &  $\frac{\sqrt{10}}{\sqrt{24}}$ \\
$\bm{M}_1^4$ & $\frac{\sqrt{18}}{\sqrt{24}}$ & $\frac{\sqrt{10}}{\sqrt{24}}$ & $\frac{\sqrt{2}}{\sqrt{24}}$  & $\frac{\sqrt{12}}{\sqrt{24}}$ &  $\frac{\sqrt{4}}{\sqrt{24}}$\\
$\bm{M}_1^5$ & $\frac{\sqrt{18}}{\sqrt{24}}$ & $\frac{\sqrt{18}}{\sqrt{24}}$ & $\frac{\sqrt{10}}{\sqrt{24}}$ & $\frac{\sqrt{16}}{\sqrt{24}}$ & $\frac{\sqrt{12}}{\sqrt{24}}$\\
\hline
\end{tabular}
\end{center}
\end{table}

From these $\tilde{d}_F(\bm{M}_1^i,\bm{M}_2^j)$ values, we get for the optimistic, pessimistic, average, and mid\footnote{i.e. Hurwicz attitude with the balance parameter $\alpha=0.5$.} attitudes 
\begin{align*}
 \tilde{d}^{\text{Optim}}(\text{Pref}_1,\text{Pref}_2)& = \min_{i,j=1,\ldots,5} \{\tilde{d}_F(\bm{M}_1^i,\bm{M}_2^j)\} = 0\\
 \tilde{d}^{\text{Pessim}}(\text{Pref}_1,\text{Pref}_2)& = \max_{i,i=1,\ldots,5} \{\tilde{d}_F(\bm{M}_1^i,\bm{M}_2^j)\}=1 \\
 \tilde{d}^{\text{Aver}}(\text{Pref}_1,\text{Pref}_2)& = \frac{1}{25} \sum_{i=1}^{5}\sum_{j=1}^{5} \tilde{d}_F(\bm{M}_1^i,\bm{M}_2^j) \approx \frac{1}{25} \cdot 17.4151  \approx 0.6966\\
 \tilde{d}^{\text{mid}}(\text{Pref}_1,\text{Pref}_2) & = \frac{1}{2} ( \tilde{d}^{\text{Optim}}(\text{Pref}_1,\text{Pref}_2) + \tilde{d}^{\text{Pessim}}(\text{Pref}_1,\text{Pref}_2))=0.5
%& = (0+ 1)/2=0.5
\end{align*}

Among all these possible decisional attitudes, we prefer using the average decision-making attitude because it exploits all possible CTPOs related to the preferences $\text{Pref}_1$ and $\text{Pref}_2$. The pessimistic, optimistic, and Hurwicz attitudes are particular cases of the average attitude based on an ad-hoc choice of weighting factors for minimum or maximum distances or both (all other weighting factors being set to zero). Based on the average attitude result, we can assert that $\text{Pref}_1$ and $\text{Pref}_2$ are more dissimilar than similar because their normalized distances are greater than 0.5. 
Based on this very simple illustrative example, we have shown how it is possible in theory to calculate the (normalized) distance between two PPOs using the brute force method. 

In practice, however, the BFM is not very simple to apply due to the very big combinatorics required, especially if the number of objects under consideration is big. The main difficulty is not computing the Frobenius distances between two CTPOs but enumerating all possible TPOs (including ties) and generating all CTPOs for each PPO. Some preliminary works on CTPOs determination problems have been addressed in \cite{Salabun2021}. Even in this very simple example, the result is definitely not so easy to obtain directly or by intuition. To circumvent the limitation of BFM and overcome its computational complexity due to combinatorics, we propose in the next section to address the problem of calculating the distance between two PPOs using a different paradigm based on belief functions.

%========================================
\section{Solution by the belief functions approach}
%========================================
\label{Sec6}

In practice, the preference ordering of two objects, say $X$ and $Y$, is not always easy to make, mainly because of the lack of pertinent/useful information to establish the scores or because of the very high complexity of calculating precisely the score values, or the difficulty to establish a personal judgment about the objects. In this situation, a choice between preferences $X\succ Y$, $X\prec Y$, and indifference $X\equiv Y$ is just impossible to make with certainty. In extreme cases, a preferential choice cannot be made at all because of the lack of evidence to support any preference. This corresponds to the full uncertain preference case denoted for short by $X?Y$ or equivalently denoted by $Y?X$. The notation $X?Y$ represents the total uncertainty about preferences $X\succ Y$, $X\prec Y$, and indifference $X\equiv Y$, which corresponds actually to their disjunction\footnote{The disjunction of two propositions $x$ and $y$ is represented by the union operation $x \cup y$ in the theory of belief functions \cite{Shafer1976}.}. More precisely, one has
\begin{equation}
X?Y=Y?X\triangleq (X\succ Y) \cup (X\equiv Y) \cup (X\prec Y).
\end{equation}

\subsection{BBA modeling from preference}
%===============================
\label{BBAmodeling}

To model the uncertainty about preference orderings, we adopt here the belief function formalism \cite{Shafer1976}, and we introduce the simplest preferential frame of discernment $\Theta_{X,Y}$ between two objects $X$ and $Y$ defined by the three exclusive and exhaustive preferences states $\theta_1\triangleq (X\succ Y)$, $\theta_2\triangleq (X\equiv Y)$, and $\theta_3\triangleq (X\prec Y)$. Hence, we will work with the FoD
\begin{equation}
\Theta_{X,Y}\triangleq \{\theta_1\triangleq X\succ Y,\theta_2\triangleq X\equiv Y,\theta_3\triangleq X\prec Y\}.
\label{FoD}
\end{equation}

The notation $\Theta_{X,Y}$ is necessary because we need to specify precisely the two elements $X$ and $Y$ involved in the analysis of the PPO under concern. We will eventually omit the $X,Y$ index for $\Theta$ whenever possible if there is no confusion on the elements $X$ and $Y$ we are working with (by default, we consider the elements or objects $X$ and $Y$).
 
To address some problems where PPOs and uncertainty occur, we can consider that the preference orderings are not necessarily true or false (as in the classical preference model), but they can better be modeled by a basic belief function
$m(.):2^{\Theta_{X,Y}} \mapsto [0,1]$ with $m(\emptyset)=0$ and $\sum_{Z\in2^{\Theta_{X,Y}}} m(Z)=1$, where $2^{\Theta_{X,Y}}$ is the power set (i.e. the set of all the subsets) of $\Theta_{X,Y}$ given by
\begin{equation*}
2^{\Theta_{X,Y}} = \{\emptyset,\theta_1,\theta_2,\theta_1\cup\theta_2,\theta_3,\theta_1\cup\theta_3,\theta_2\cup\theta_3, \theta_1\cup\theta_2\cup\theta_3 \}.
\end{equation*}
The disjunctions $\theta_1\cup\theta_2$, $\theta_1\cup\theta_3$, $\theta_2\cup\theta_3$ and $\theta_1\cup\theta_2\cup\theta_3$ correspond to the following preference situations:
\begin{align*}
& (X\succeq Y)\triangleq \theta_1\cup\theta_2= (X\succ Y) \cup (X\equiv Y)\\
& (X\not\equiv Y)\triangleq \theta_1\cup\theta_3= (X\succ Y) \cup (X\prec Y)\\
& (X\preceq Y)\triangleq \theta_2\cup\theta_3= (X\equiv Y)\cup (X\prec Y)\\
& (X?Y)\triangleq \theta_1\cup \theta_2\cup\theta_3= (X\succ Y) \cup (X\equiv Y) \cup (X\prec Y)
\end{align*}

The use of a mass function $m(.)$ defined on the FoD $\Theta_{X,Y}$ offers the great advantage of characterizing any type of PO (including probabilistic preference orderings and the total uncertain preference $X?Y$) as shown below:
\begin{enumerate}
\item {\bf{Case 1}}: This corresponds to the classical case where $X\succ Y$ is declared (or chosen) for sure. For this case, we take ${m(\theta_1)=m(X\succ Y)=1}$, all other mass values equal zero, and we get the belief intervals $[Bel(X\succ Y),Pl(X\succ Y)]=[1,1]$,  $[Bel(X\equiv Y),Pl(X\equiv Y)]=[0,0]$, and $[Bel(X\prec Y),Pl(X\prec Y)]=[0,0]$. 
\item {\bf{Case 2}}: This corresponds to the classical case where $X\equiv Y$ is declared or chosen for sure. We take ${m(\theta_2)=m(X\equiv Y)=1}$, all other mass values equal zero, and we get the belief intervals $[Bel(X\equiv Y),Pl(X\equiv Y)]=[1,1]$,  $[Bel(X\succ Y),Pl(X\succ Y)]=[0,0]$, and $[Bel(X\prec Y),Pl(X\prec Y)]=[0,0]$. 
\item {\bf{Case 3}}: This corresponds to the classical case where $X\prec Y$ is declared or chosen for sure. We take ${m(\theta_3)=m(X\prec Y)=1}$, all other mass values equal zero, and we get the belief intervals $[Bel(X\prec Y),Pl(X\prec Y)]=[1,1]$,  $[Bel(X\succ Y),Pl(X\succ Y)]=[0,0]$, and $[Bel(X\equiv Y),Pl(X\equiv Y)]=[0,0]$. 
\item  {\bf{Case 4}}: This corresponds to the vacuous (i.e. the total ignorant) case $X?Y$ for sure when there is no useful evidence to help to select $X\succ Y$, $X\equiv Y$, or $X\prec Y$. We take $m(\theta_1\cup\theta_2\cup \theta_3)=m((X\succ Y)\cup(X\equiv Y)\cup(X\prec Y))=1$, all other mass values equal zero, and we get the belief intervals $[Bel(X\succ Y),Pl(X\succ Y)]=[0,1]$,  $[Bel(X\equiv Y),Pl(X\equiv Y)]=[0,1]$, and $[Bel(X\prec Y),Pl(X\prec Y)]=[0,1]$. 
\item  {\bf{Case 5}}: This is the Bayesian (i.e., probabilistic) BBA case where we put a positive mass only to singletons of the power set $2^{\Theta_{X,Y}}$. For instance, we could take $m(\theta_1)=m(X\succ Y)=0.2$, $m(\theta_2)=m(X\equiv Y)=0.3$ and $m(\theta_3)=m(X\prec Y)=0.5$, all other mass values equal zero, and we would get the belief intervals $[Bel(X\succ Y),Pl(Y\succ Y)]=[0.2,0.2]$,  $[Bel(X\equiv Y),Pl(X\equiv Y)]=[0.3,0.3]$, and $[Bel(X\prec Y),Pl(X\prec Y)]=[0.5,0.5]$, which represents the (subjective) probabilities $P(X\succ Y)=0.2$, $P(X\equiv Y)=0.3$ and $P(X\prec Y)=0.5$ of possible preference states.
\item  {\bf{Case 6}}: This is the general case modeling imprecise preference where all the elements of $2^{\Theta_{X,Y}}\setminus\{\emptyset\}$ are focal elements of $m(.)$. For instance, $m(\theta_1)=0.1$, $m(\theta_2)=0.2$, $m(\theta_3)=0.3$, $m(\theta_1\cup \theta_2)=0.02$, $m(\theta_1\cup \theta_3)=0.03$, $m(\theta_2\cup \theta_3)=0.05$, $m(\theta_1\cup \theta_2\cup \theta_3)=0.3$. Based on this BBA, we get the belief intervals $[Bel(X\succ Y),Pl(X\succ Y)]=[0.1,0.45]$,  $[Bel(X\equiv Y),Pl(X\equiv Y)]=[0.2,0.57]$, and $[Bel(X\prec Y),Pl(X\prec Y)]=[0.3,0.68]$, which represent imprecise probabilities $P(X\succ Y)\in [0.1,0.45]$, $P(X\equiv Y)\in [0.2,0.57]$ and $P(X\prec Y)\in [0.3,0.68]$ of the preference states.
\end{enumerate}

\subsection{Construction of BBA matrices based on belief masses}
%=================================================

Because it is impossible to calculate each component $M(i,j)$ of PSM from \eqref{eqPSM} without high combinatorics (see BFM of Section \ref{Sec5}), we propose to circumvent this combinatorial problem by exploiting belief functions. Before calculating the distance between PPOs based on the belief functions, it is necessary to construct a BBA matrix associated with each (partial) preference ordering based on the BBA modeling from preferences presented just before. Two types of methods are analyzed in this work to establish the distance between two PPOs: the direct method based on the BBAs matrices, which requires calculating the Frobenius distance between two (${8N\times 8N}$) matrices\footnote{$N$ being the cardinality of the set of objects under concern.}, and two indirect methods based on distances between BBAs, which require the calculation of the Frobenius distance between two PSM-alike matrices of dimension $N\times N$ only. We present briefly these different approaches next.

\subsubsection{Direct method}
%=======================

For each object $X_i$ (${i=1,\ldots,N}$) we first establish the BBA matrices $\textbf{BBA}_1$ and  $\textbf{BBA}_2$ from preferences ${\text{Pref}_1}$ and ${\text{Pref}_2}$ respectively based on the previous BBA modeling (see subsection \ref{BBAmodeling}, and the next examples). Because we work with FoD of cardinality three defined by \eqref{FoD}, each BBA of the BBA matrix has $2^3=8$ components. Hence, ${\textbf{BBA}_1}$ and  ${\textbf{BBA}_2}$ matrices will be sparse matrices of dimension ${8N\times 8N}$. Once the $\textbf{BBA}_1$ and  $\textbf{BBA}_2$ are built, we use the \enquote{extended PSM} matrices ${\bm{M}_1=  \textbf{BBA}_1}$ and ${\bm{M}_2=  \textbf{BBA}_2}$. Then, we apply the classical Frobenius distance definition ${d_F(\bm{M}_1,\bm{M}_2)}$ to calculate the distance between the two PPOs ${\text{Pref}_1}$ and ${\text{Pref}_2}$. It is worth mentioning that in order to deal with the mathematical representation of uncertainty by belief functions, we have changed the paradigm, and the \enquote{PSM matrices} are no longer ${N\times N}$ matrices anymore (as defined classically) but extended into ${8N\times 8N}$ BBA matrices whose components correspond to belief mass values defined with respect to different FoD defined in \eqref{FoD}.

\subsubsection{Indirect methods}
%========================
In order to work with belief functions and with PSM matrices of dimension ${N\times N}$ instead of ${8N\times 8N}$, we examine the possibility to redefine each component $M(i,j)$ of PSM-alike matrix from the distance of the BBA $m_{\Theta_{X_i,X_j}}$ with respect to the BBA $m_{X_i>X_j}$. Here $m_{\Theta_{X_i,X_j}}$ is the BBA built on ${\Theta_{X_i,X_j}}$ from the given (partial) preference ordering, and $m_{X_i>X_j}$ is the BBA focused on the element ${\theta_1=X_i\succ X_j}$ of the FoD $\Theta_{X_i,X_j}$ (i.e. ${m_{X_i>X_j}(\theta_1)=1}$). More precisely, we will define the Preference-Score alike Matrix $\bm{M}(X)=[{M}(i,j)]$ from its components ${M}(i,j)$ for $i,j=1,2,\ldots,n$ defined as
\begin{equation}
M(i,j)\triangleq d(m_{\Theta_{X_i,X_j}},m_{X_i>X_j}).
\label{eqPSMBF}
\end{equation}

where $d(\cdot,\cdot)$ is a distance measure between two BBAs. In this paper, we use Jousselme distance $d_J$ and the belief-interval-based distance $d_{BI}$ to get results because they are the two main well-known normalized distances often used in the BF framework. Due to space limit restriction, we do not give definitions of these distances, which are given in \cite{Jousselme2001,Jousselme2012} for $d_J$, and in \cite{Dezert2016,Han2018} for $d_{BI}$. Once the PSMs-alike are obtained by \eqref{eqPSMBF}, we use the classical Frobenius distance definition \eqref{FrobeniusDistance} to calculate the distance between the two PPOs.

We analyze how the direct and indirect methods exploiting belief functions work in examples 1 \& 2 to justify our choice of the direct method with respect to the indirect methods.

\subsection{Example 1 revisited}
%========================

We present the results of the direct and indirect methods using the belief function, for the example, 1 of Section \ref{Sec3} involving 3 objects $\{X_1\triangleq A, X_2\triangleq B, X_3\triangleq C\}$ and the two TPOs ${\text{Pref}_1\triangleq B \succ A \succ C}$ and ${\text{Pref}_2\triangleq B\succ C \succ A}$. The matrix $\textbf{BBA}_1$ of \eqref{eqBBA1} requires the BBAs $\bm{m}_{i,j}$ ($i,j=1,2,3$) with the following masses of focal elements:
${m_{1,2}(\theta_3=(X_1\prec X_2)=1}$, ${m_{1,3}(\theta_1=(X_1\succ X_3))=1}$, 
${m_{2,1}(\theta_1=(X_2\succ X_1)=1}$, ${m_{2,3}(\theta_1=(X_2\succ X_3))=1}$, 
${m_{3,1}(\theta_3=(X_3\prec X_1))=1}$, and ${m_{3,2}(\theta_3=(X_3\prec X_2))=1}$.  For the diagonal BBAs, 
${m_{1,1}(\theta_2=(X_1\equiv X_1))=1}$, ${m_{2,2}(\theta_2=(X_2\equiv X_2))=1}$, and ${m_{3,3}(\theta_2=(X_3\equiv X_3))=1}$.
Therefore, we have\footnote{For simplicity, we denote $\bm{m}_{\Theta_{X_i,X_j}}$ as $\bm{m}_{i,j}$ for $i,j=1,2,3$.}
\begin{equation}
\textbf{BBA}_1 \triangleq
\begin{bmatrix}
\bm{m}_{1,1} & \bm{m}_{1,2} & \bm{m}_{1,3} \cr
\bm{m}_{2,1} & \bm{m}_{2,2} & \bm{m}_{2,3} \cr
\bm{m}_{3,1} & \bm{m}_{3,2} & \bm{m}_{3,3}\cr	
\end{bmatrix}
\label{eqBBA1}
\end{equation}
with mass functions defined by the vectors\footnote{The components of mass vectors $\bm{m}_{i,j}$ (from the left to the right) correspond respectively to $m_{i,j}=(\emptyset)$, $m_{i,j}=(\theta_1)$, $m_{i,j}=(\theta_2)$, $m_{i,j}=(\theta_1\cup\theta_2)$, $m_{i,j}=(\theta_3)$, $m_{i,j}=(\theta_1\cup\theta_3)$, $m_{i,j}=(\theta_2\cup\theta_3)$ and $m_{i,j}=(\theta_1\cup\theta_2\cup\theta_3)$.}
\begin{align*}
 \bm{m}_{1,1} = [0\, 0\, 1\, 0\,&0\, 0\, 0\, 0],\ \bm{m}_{1,2} =[0\, 0\, 0\, 0\, 1\, 0\, 0\, 0], \bm{m}_{1,3}= [0\, 1\, 0\, 0\, 0\, 0\, 0\, 0]\\
 \bm{m}_{2,1} = [0\, 1\, 0\, 0\,&0\, 0\, 0\, 0],\ \bm{m}_{2,2} =[0\, 0\, 1\, 0\, 0\, 0\, 0\, 0], \bm{m}_{2,3}= [0\, 1\, 0\, 0\, 0\, 0\, 0\, 0]\\
 \bm{m}_{3,1} = [0\, 0\, 0\, 0\,&1\, 0\, 0\, 0],\ \bm{m}_{3,2} =[0\, 0\, 0\, 0\, 1\, 0\, 0\, 0], \bm{m}_{3,3}= [0\, 0\, 1\, 0\, 0\, 0\, 0\, 0]
\end{align*}
Similarly, we will have
\begin{equation}
\textbf{BBA}_2 \triangleq
\begin{bmatrix}
\bm{m}_{1,1} & \bm{m}_{1,2} & \bm{m}_{1,3} \cr
\bm{m}_{2,1} & \bm{m}_{2,2} & \bm{m}_{2,3} \cr
\bm{m}_{3,1} & \bm{m}_{3,2} & \bm{m}_{3,3}\cr	
\end{bmatrix}
\label{eqBBA2}
\end{equation}
with mass functions defined by the vectors
\begin{align*}
\bm{m}_{1,1} = [0\, 0\, 1\, 0\, 0\,&0\, 0\, 0],\ \bm{m}_{1,2} =[0\, 0\, 0\, 0\, 1\, 0\, 0\, 0], \bm{m}_{1,3}= [0\, 0\, 0\, 0\, 1\, 0\, 0\, 0]\\
\bm{m}_{2,1} = [0\, 1\, 0\, 0\, 0\,&0\, 0\, 0],\ \bm{m}_{2,2} =[0\, 0\, 1\, 0\, 0\, 0\, 0\, 0], \bm{m}_{2,3}= [0\, 1\, 0\, 0\, 0\, 0\, 0\, 0]\\
\bm{m}_{3,1} = [0\, 1\, 0\, 0\, 0\,&0\, 0\, 0],\ \bm{m}_{3,2} =[0\, 0\, 0\, 0\, 1\, 0\, 0\, 0], \bm{m}_{3,3}= [0\, 0\, 1\, 0\, 0\, 0\, 0\, 0]
\end{align*}

With the direct method, we take ${\bm{M}_1=  \textbf{BBA}_1}$ and ${\bm{M}_2=  \textbf{BBA}_2}$, and therefore the distance between ${\text{Pref}_1\triangleq B \succ A \succ C}$ and ${\text{Pref}_2\triangleq B\succ C \succ A}$ is $d_F(\bm{M}_1,\bm{M}_2)=2$. It can be easily verified\footnote{By considering two full conflicting total preference orderings, say ${\text{Pref}_1\triangleq B \succ A \succ C}$ and ${\text{Pref}_3\triangleq C \succ A \succ B}$.}  that the maximal Frobenius distance is $d_F(\bm{M}_1,\bm{M}_2)\approx 3.4641$. Therefore, the normalized distance between ${\text{Pref}_1}$ and ${\text{Pref}_2}$ is $\tilde{d}_F(\bm{M}_1,\bm{M}_2)=2/3.4641=0.5774$. This result coincides with the normalized Frobenius distance given by the classical method described in the subsection \ref{Example1}.\medskip

With the indirect methods, it turns out that we obtain the same PSM-alike matrices ${\bm{M}_1}$ and ${\bm{M}_2}$ if we use Jousselme distance \cite{Jousselme2001}, or the belief-interval based distance  \cite{Dezert2016}, and we get
%
%{\small{
%\begin{equation*}
%{\bm{M}_1}  =
%\begin{bmatrix}
%1 & 1 & 0  \cr
%0 & 1 & 0  \cr
% 1 & 1 & 1	
%\end{bmatrix},	
%{\bm{M}_2}  =
%\begin{bmatrix}
% 1 & 1 & 1  \cr
%0 & 1 & 0  \cr
%0 & 1 & 1 \cr		
%\end{bmatrix}.						
%\end{equation*}
%}}

\begin{equation*}
{\bm{M}_1}  =
\begin{bmatrix}
1 & 1 & 0  \cr
0 & 1 & 0  \cr
 1 & 1 & 1	
\end{bmatrix},	
\quad \text{and}\quad
{\bm{M}_2}  =
\begin{bmatrix}
 1 & 1 & 1  \cr
0 & 1 & 0  \cr
0 & 1 & 1 \cr		
\end{bmatrix}.						
\end{equation*}

Therefore, the Frobenius distance defined by \eqref{FrobeniusDistance} based on PSMs-alike built from Jousselme distance gives
 $${d}^{\text{J}}(\text{Pref}_1,\text{Pref}_2) = {d}^{\text{BI}}(\text{Pref}_1,\text{Pref}_2) \approx 1.4142$$
Because ${d}^{\text{J}}_{\max}={d}^{\text{BI}}_{\max}\approx 2.4495$, we get finally the normalized distance between ${\text{Pref}_1}$ and ${\text{Pref}_2}$
 $$\tilde{d}^{\text{J}}(\text{Pref}_1,\text{Pref}_2) = \tilde{d}^{\text{BI}}(\text{Pref}_1,\text{Pref}_2) \approx 1.4142/2.4495 = 0.5774.$$
We note that this value 0.5774 based on belief functions and Jousselme distance coincides with the normalized distance we obtain by the direct method and the classical method described in the subsection \ref{Example1}.

So, we see that the result of our direct and indirect methods based on belief functions for calculating the distance between two total preference orderings coincide with the classical method, which is normal because there is no uncertainty involved in such a case.

\subsection{Example 2 revisited}
%========================

Consider 3 objects $\{X_1\triangleq A, X_2\triangleq B, X_3\triangleq C\}$ and the two PPOs $ \text{Pref}_1 = C \succ A$, and $\text{Pref}_2 = A\succ B$.
From the knowledge of $\text{Pref}_1$, the BBA matrix $\textbf{BBA}_1$ of the form \eqref{eqBBA1} has the following components
\begin{align*}
\bm{m}_{1,1} = [0\, 0\, 1\, 0\,&0\, 0\, 0\, 0],\ \bm{m}_{1,2} =[0\, 0\, 0\, 0\, 0\, 0\, 0\, 1], \bm{m}_{1,3}= [0\, 0\, 0\, 0\, 1\, 0\, 0\, 0]\\
\bm{m}_{2,1} = [0\, 0\, 0\, 0\,&0\, 0\, 0\, 1],\ \bm{m}_{2,2} =[0\, 0\, 1\, 0\, 0\, 0\, 0\, 0], \bm{m}_{2,3}= [0\, 0\, 0\, 0\, 0\, 0\, 0\, 1]\\
\bm{m}_{3,1} = [0\, 1\, 0\, 0\,&0\, 0\, 0\, 0],\ \bm{m}_{3,2} =[0\, 0\, 0\, 0\, 0\, 0\, 0\, 1], \bm{m}_{3,3}= [0\, 0\, 1\, 0\, 0\, 0\, 0\, 0]
\end{align*}

The BBAs $\bm{m}_{1,2}$, $\bm{m}_{2,1}$, $\bm{m}_{2,3}$, and $\bm{m}_{3,2}$ correspond to the vacuous belief masses because we do not have information about the preference between $B$ and $A$ and between $B$ and  $C$ based on the PPO $ \text{Pref}_1 = C \succ A$.

Similarly, from the knowledge of $\text{Pref}_2$, the BBA matrix $\textbf{BBA}_2$ of the form \eqref{eqBBA2} has the following components
\begin{align*}
\bm{m}_{1,1} = [0\, 0\, 1\, 0\,&0\, 0\, 0\, 0],\ \bm{m}_{1,2} =[0\, 1\, 0\, 0\, 0\, 0\, 0\, 0], \bm{m}_{1,3}= [0\, 0\, 0\, 0\, 0\, 0\, 0\, 1]\\
\bm{m}_{2,1} = [0\, 0\, 0\, 0\,&1\, 0\, 0\, 0],\ \bm{m}_{2,2} =[0\, 0\, 1\, 0\, 0\, 0\, 0\, 0], \bm{m}_{2,3}= [0\, 0\, 0\, 0\, 0\, 0\, 0\, 1]\\
\bm{m}_{3,1} = [0\, 0\, 0\, 0\,&0\, 0\, 0\, 1],\ \bm{m}_{3,2} =[0\, 0\, 0\, 0\, 0\, 0\, 0\, 1], \bm{m}_{3,3}= [0\, 0\, 1\, 0\, 0\, 0\, 0\, 0]
\end{align*}

With the direct method, we take $\bm{M}_1=  \textbf{BBA}_1$ and $\bm{M}_2=  \textbf{BBA}_2$, and therefore the distance between $\text{Pref}_1\triangleq C \succ A$ and $\text{Pref}_2\triangleq A\succ B$ is ${d_F(\bm{M}_1,\bm{M}_2)=2.8284}$. Because the maximal Frobenius distance is ${d_F(\bm{M}_1,\bm{M}_2)\approx 3.4641}$, the normalized distance between ${\text{Pref}_1}$ and ${\text{Pref}_2}$ is $\tilde{d}_F(\bm{M}_1,\bm{M}_2)=2.8284/3.4641=0.8165$. This distance is bigger than the average-based distance $\tilde{d}^{\text{Aver}}(\text{Pref}_1,\text{Pref}_2) = 0.6966$, and the mid-based distance $\tilde{d}^{\text{mid}}(\text{Pref}_1,\text{Pref}_2) = 0.5$ obtained by the brute force method (see subsection \ref{Example 2}). This result reinforces the conclusion that $\text{Pref}_1\triangleq C \succ A$  is more dissimilar to $\text{Pref}_2\triangleq A\succ B$ than similar to $\text{Pref}_2$ because $\tilde{d}_F(\bm{M}_1,\bm{M}_2)$ is much greater than 0.5. The advantage of this direct belief-based approach is that its conclusion is coherent with the conclusion drawn from the average attitude obtained by the brute force approach without the need for high combinatorics. This direct method has a quadratic complexity of $64 N^2$.\medskip

With the indirect method using Jousselme distance \cite{Jousselme2001} we get
\begin{equation*}
{\bm{M}_1}  =
\begin{bmatrix}
1 & 0.8165 & 1  \cr
0.8165 & 1 & 0.8165 \cr
 0 & 0.8165 & 1	
\end{bmatrix},
\quad \text{and}\quad
{\bm{M}_2}  =
\begin{bmatrix}
 1 & 0 & 0.8165  \cr
1 & 1 & 0.8165  \cr
0.8165 & 0.8165 & 1 \cr		
\end{bmatrix}.	
%\label{eqMX2BF}						
\end{equation*}

Therefore, the Frobenius distance defined by \eqref{FrobeniusDistance} is ${d}^{\text{J}}(\text{Pref}_1,\text{Pref}_2)\approx 1.1835$. 
Because ${d}^{\text{J}}_{\max}\approx 2.4495$, we get finally the normalized distance between ${\text{Pref}_1}$ and ${\text{Pref}_2}$
 $$\tilde{d}^{\text{J}}(\text{Pref}_1,\text{Pref}_2) \approx 1.1835/2.4495 = 0.4832.$$
This distance value of 0.4832 is a little bit lower than $\tilde{d}^{\text{mid}}(\text{Pref}_1,\text{Pref}_2) = 0.5$ obtained by the direct method above, and the brute force method based on mid attitude and on average attitude described in the subsection \ref{Example 2}. Hence, the conclusion drawn is now different from the conclusion drawn from the previous direct method because we will conclude that $\text{Pref}_1$ is a bit similar to $\text{Pref}_2$ because $\tilde{d}^{\text{J}}(\text{Pref}_1,\text{Pref}_2)$ is a bit lower than 0.5.\medskip

With the indirect method using belief-interval based distance \cite{Dezert2016} we get
\begin{equation*}
{\bm{M}_1}  =
\begin{bmatrix}
1 & 0.7071 & 1  \cr
0.7071 & 1 & 0.7071 \cr
 0 & 0.7071 & 1	
\end{bmatrix},	
\quad \text{and}\quad
{\bm{M}_2}  =
\begin{bmatrix}
 1 & 0 & 0.7071  \cr
1 & 1 & 0.7071  \cr
0.7071 & 0.7071 & 1 \cr		
\end{bmatrix}.						
\end{equation*}

Therefore, the Frobenius distance defined by \eqref{FrobeniusDistance} is
 $${d}^{\text{BI}}(\text{Pref}_1,\text{Pref}_2) \approx 1.0824$$
Because ${d}^{\text{BI}}_{\max}\approx 2.4495$, we get finally the normalized distance between ${\text{Pref}_1}$ and ${\text{Pref}_2}$
 $$\tilde{d}^{\text{BI}}(\text{Pref}_1,\text{Pref}_2) \approx 1.0824/2.4495 = 0.4419.$$
This normalized distance value 0.4419 is lower than 0.5 and lower than $\tilde{d}_F(\bm{M}_1,\bm{M}_2)$ obtained by the direct method and the brute force method based on mid attitude and on average attitude described in the subsection \ref{Example 2}. It is also lower than $\tilde{d}^{\text{J}}(\text{Pref}_1,\text{Pref}_2) \approx 0.4832.$, and thus it reinforces the conclusion that $\text{Pref}_1$ is more similar to $\text{Pref}_2$ than dissimilar to it because $\tilde{d}^{\text{BI}}(\text{Pref}_1,\text{Pref}_2)$ is lower than 0.5.

This example involving two PPOs shows clearly that very different conclusions can be drawn by the direct method and indirect methods exploiting belief functions. These different results and conclusions are not so surprising, and they can be explained by the fact that the indirect methods make a kind of \enquote{lossy compression} of information by using ${N\times N}$ PSM-alike matrices, whereas the direct method works with extended matrices of dimension ${8N\times 8N}$ characterizes the whole information available for modeling adequately the uncertainty inherent to the PPOs. So we think that it is risky to use indirect methods in general for measuring the distance between two PPOs, and we recommend using the direct method even if it requires working with two matrices of size ${8N\times 8N}$ instead of working with two ${N\times N}$ matrices. It should be noted that indirect methods also require an ad hoc choice of metric (Jousselme's distance, belief interval distance, etc.), which could also be the subject of debate.

%=============================================
\section{Conclusion and perspectives}
%=============================================
\label{Sec7}

In this paper, we have shown how it is possible to measure the distance between two PPOs either by a brute force method or by methods using belief functions that can properly model the uncertainty inherent in PPOs. The brute force method becomes intractable in practical problems involving many objects in PPOs due to the combinatorial complexity, and this will limit its use in practice. The direct method based on belief functions is attractive because it requires only two sparse matrices of size $64N^2$, which are easy to construct, and it does not require combinatorics, which is a serious advantage over the brute force method for dealing with high-dimensional problems. Our new direct method can also deal with any kind of basic belief assignments (not necessarily with mass values set to 0 or 1 as used in this work) to represent possibly imprecise preferences, if one needs to deal with them. As research perspectives, it would be interesting to see if it is possible to extend this approach to calculate the distance between \enquote{hybrid} partial preference orderings (HPPOs) involving disjunctions or conjunctions of PPOs, for instance between preferences expressed as ${\text{Pref}_1=(A\succ D)\vee (C\succ B\succ E)}$ and ${\text{Pref}_2=(A\succ B \succ E)\wedge (D\succ C)}$. This theoretical work provides a new tool for dealing with partial preference orders, which could hopefully help in the development of new fusion techniques for multi-criteria decision support.

%========================================================================================================


\begin{thebibliography}{99}
%========================================================================================================


\bibitem{Critchlow1985}
D.E.~Critchlow, \emph{Metric methods for analyzing partially ranked data}, Lecture Notes in Statistics, Springer, 1985.

\bibitem{Fagin2004}
R.~Fagin, R.~Kumar, M.~Mahdian, D.~Sivakumar, E.~Vee, \emph{Comparing and aggregating rankings with ties}, Proc. of the 23th ACM SIGMOD-SIGACT-SIGART Symposium on Principles of database systems, pp.~47--58, 2004.

\bibitem{Fagin2006}
R.~Fagin, R.~Kumar, M.~Mahdian, D.~Sivakumar, E.~Vee, \emph{Comparing partial rankings}, SIAM Journal on Discrete Mathematics, Vol.~20, No.~3, pp.~628--648, 2006.

\bibitem{draws2023}
W.~Sa\l{}abun, A.~Shekhovtsov, \emph{An Innovative Drastic Metric for Ranking Similarity in Decision-Making Problems},  Proc. of the 18th Conference on Computer Science and Intelligence Systems (FedCSIS), pp.~731--738, 2023.

\bibitem{diaconis1977spearman}
P.~Diaconis, R.L.~Graham, \emph{Spearman's footrule as a measure of disarray}, Journal of the Royal Statistical Society Series B: Statistical Methodology, Vol.~39, No.~2, pp.~262--268, 1977.
% Oxford University Press

\bibitem{Bansal2009}
M.S.~Bansal, D.~Fern{\'a}ndez-Baca, \emph{Computing distances between partial rankings}, Information Processing Letters, Vol.~109, No.~4, pp.~238--241, 2009.
% Elsevier

\bibitem{Albano2021}
A.~Albano, A.~Plaia, \emph{Element weighted Kemeny distance for ranking data}, Electronic Journal of Applied Statistical Analysis, Vol.~14, No.~1, pp.~117--145, 2021.

\bibitem{Kumar2010}
R.~Kumar, S.~Vassilvitskii, \emph{Generalized distances between rankings}, Proc. of the 19th International Conference on World Wide Web, pp.~571--580, 2010.

\bibitem{Kemeny1962}
J.L.~Snell, J.G.~Kemeny, \emph{Mathematical models in the social sciences}, 1962.

\bibitem{Frobenius2023}
J.~Dezert, A.~Shekhovtsov, W.~Sa\l{}abun, \emph{A new distance between rankings}, Heliyon, Vol.~10, No.~7, 2024.
% Elsevier



\bibitem{Horn1990}
R.A.~Horn, C.R.~Johnson, \emph{Matrix analysis}, Cambridge University Press, 2012.

\bibitem{Golub1996}
G.H.~Golub, C.F.~Van Loan, \emph{Matrix computations}, JHU Press, 2013.


\bibitem{Shafer1976}
G.~Shafer, \emph{A mathematical theory of evidence}, Princeton University Press, 1976.
% Volume 42

\bibitem{Hurwicz1951}
L.~Hurwicz, \emph{The generalized Bayes minimax principle: a criterion for decision making under uncertainty}, Cowles Comm. Discuss. Paper Stat., Vol.~335, p.~1950, 1951.

\bibitem{Hurwicz1952}
L.~Hurwicz, \emph{A criterion for decision making under uncertainty}, Technical Report 355, Cowles Commission, 1952.

\bibitem{Salabun2021}
W.~Sa\l{}abun, A.~Shekhovtsov, B.~Kizielewicz, \emph{A new consistency coefficient in the multi-criteria decision analysis domain}, Proc. of Int. Conf. on Computational Science, pp.~715--727, 2021.
%Springer

\bibitem{Jousselme2001}
A.-L.~Jousselme, D.~Grenier, {\'E}.~Boss{\'e}, \emph{A new distance between two bodies of evidence}, Information fusion, Vol.~2, No.~2, pp.~91--101, 2001.
% Elsevier

\bibitem{Jousselme2012}
A.-L.~Jousselme, P.~Maupin, \emph{Distances in evidence theory: Comprehensive survey and generalizations}, International Journal of Approximate Reasoning, Vol.~53, No.~2, pp.~118--145, % Elsevier

\bibitem{Dezert2016}
J.~Dezert, D.~Han, J.-M.~Tacnet, S.~Carladous, Y.~Yang, \emph{Decision-making with belief interval distance}, Proc. of Belief Functions: Theory and Applications: 4th International Conference, BELIEF 2016, pp.~66--74, Prague, Czech Republic, September 21--23, 2016.
%Springer

\bibitem{Han2018}
D.~Han, J.~Dezert, Y.~Yang, \emph{Belief interval-based distance measures in the theory of belief functions}, IEEE Trans. on Systems, Man, and Cybernetics: Systems, Vol.~48, No.~6, pp.~833--850, 2018.



\end{thebibliography}
\end{document}